\documentclass[letterpaper, 10 pt, conference]{ieeeconf}  

\usepackage{siunitx}
\usepackage{graphicx} 
\usepackage{makecell}
\usepackage{lipsum}

\usepackage{amsmath,amssymb,amsthm}
\usepackage{ upgreek }
\usepackage{booktabs}
\usepackage{multirow}
\usepackage{tabularx}
\usepackage{rotating}
\usepackage{bigstrut}
\newcommand{\duy}{0mm}
\usepackage{color, colortbl}

\IEEEoverridecommandlockouts                              

\usepackage[maxfloats=256]{morefloats}
\makeatletter
\let\NAT@parse\undefined
\makeatother
\usepackage[bookmarks=false, linkcolor=blue, urlcolor=blue, citecolor=blue]{hyperref} 
\usepackage{fancyhdr}
\usepackage{cleveref}
\hypersetup{
    colorlinks=true,
    linkcolor=red,
    filecolor=magenta,      
    urlcolor=blue,
    pdfstartview={FitH},
    citecolor =blue
    }
\title{\LARGE \bf
Probe-to-Grasp Manipulation Using Self-Sensing\\Pneumatic Variable-Stiffness Joints}

\author{Ngoc Duy Tran, Yeman Fan, Feng Dai, Khang Nguyen, Anh Nguyen, Hoang Hiep Ly, Tung D. Ta, Shigeru Chiba
}

\begin{document}
\maketitle
\thispagestyle{empty}
\pagestyle{empty}

\begin{abstract}
Grasping deformable objects with varying stiffness remains a significant challenge in robotics. Estimating the local stiffness of a target object is important for determining an optimal grasp pose that enables stable pickup without damaging the object. This paper presents a probe-to-grasp manipulation framework for estimating the relative stiffness of objects using a passive soft-rigid two-finger hybrid gripper equipped with self-sensing pneumatic variable-stiffness joints. Each finger of the gripper consists of two rigid links connected by a soft pneumatic ring placed at the joint, enabling both compliant interaction and controllable joint stiffness via internal pressurization. By measuring the pressure inside the pneumatic ring, we can estimate the interaction force during contact. Building on this, we propose a practical probing strategy to infer relative object stiffness by correlating the estimated normal force with known gripper closing displacement. We validate the self-sensing model through stiffness characterization experiments across bending angles and pressure ranges, and demonstrate stiffness-aware probing-and-grasping in real-life applications: selecting grasp locations on fruits with spatially varying stiffness. The proposed system offers a minimal, low-cost sensing approach for stiffness-aware soft manipulation while retaining probing and grasping capability.

\end{abstract}

\section{INTRODUCTION}






% ...

Soft robotic grippers have emerged as a promising solution for safe and adaptive manipulation, particularly when interacting with delicate and deformable objects \cite{shintake2018soft}. Unlike rigid grippers, which apply relatively fixed and concentrated contact forces that may damage soft items, soft grippers inherently distribute and modulate contact forces through structural compliance. This variable-force characteristic makes them especially suitable for handling fragile objects in unstructured environments \cite{shintake2018soft, fan2024overview}. However, while soft grippers can generate sufficient grasping force, accurately measuring and controlling the contact force exerted on the object remains a significant challenge. Sensory feedback is therefore essential to fully exploit the advantages of soft manipulation.

Existing approaches to force estimation in soft grippers generally rely on embedded force sensors or specialized smart materials. These include piezoresistive, capacitive, and magnetic elastomer-based sensing elements integrated into the soft structure \cite{yang2025soft, li2026labor, ji2025multisensory, liu2024three}. Although effective, such solutions often increase system complexity, cost, and fabrication difficulty. Alternatively, vision-based approaches, such as marker tracking, motion capture systems, or deformation-based inference using mechanical models, have been proposed to estimate contact forces indirectly. These methods reduce hardware complexity but may require sophisticated setups or calibration procedures.

In this work, we investigate a minimal and easily deployable approach to contact force and stiffness estimation using a pressure sensor. By tracking the change in value of pressure inside a soft ring incorporating a measured look-up data table, we estimate the interaction force without additional embedded force sensors. This setup requires only a few simple installations, significantly lowering the cost of implementation compared to conventional tactile sensing systems.


Based on the contact force estimated by the change of pressure, we further propose a strategy to infer the stiffness of the grasped object. By correlating interaction force and deformation during grasping, the system can approximate object mechanical properties without dedicated tactile arrays or complex instrumentation.

The main contributions of this paper are as follows:
\begin{enumerate}
    \item Proposing a passive pneumatic hybrid gripper system that enables both probing and grasping with a simple pressure sensor.
    
    \item Introducing a practical strategy for estimating object stiffness based on the adjustable stiffness capability and intrinsic parameters of the pneumatic hybrid gripper.
\end{enumerate}

    


    \begin{figure}
        \centering
        \includegraphics[width=1\linewidth]{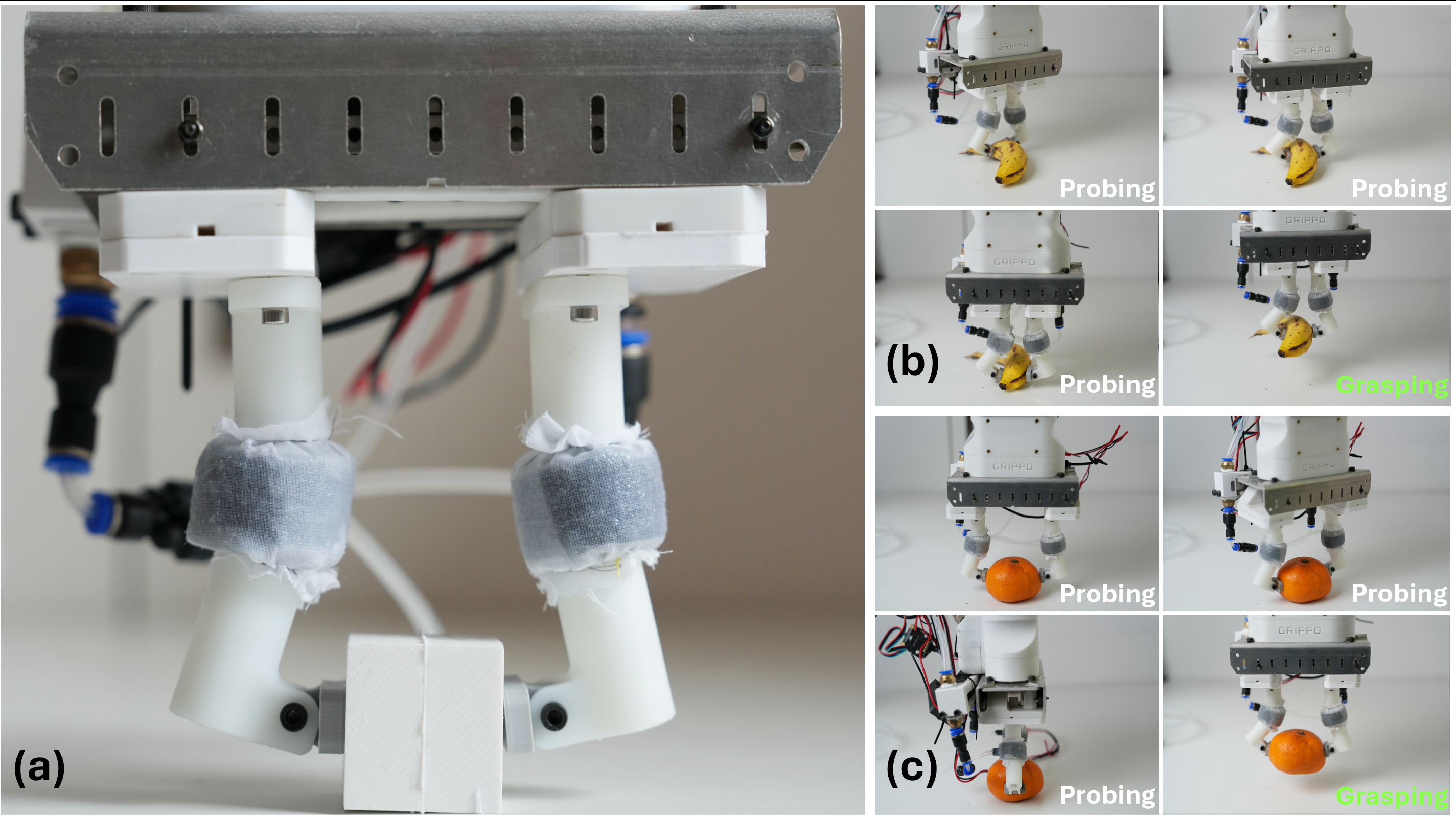}
        \caption{Overview of the hybrid gripper mounted on a robotic arm during grasping and probing tasks. (a) The gripper integrates rigid links with a pneumatic soft-ring joint, enabling compliant interaction and variable stiffness control. (b) The hybrid gripper performs a probing process on elongated fruits, such as a banana, to estimate stiffness before grasping. (c) For round fruits such as an orange, the gripper rotates around the object to probe different locations and identify the stiffest region for a stable grasp.}
        \label{fig:placeholder}
        \vspace*{\duy}
    \end{figure}

\section{RELATED WORK} \label{sec:background}

Estimating object stiffness is an important problem in robotics and haptics \cite{chin2020machine}. Existing approaches can gener-
ally be categorized into absolute stiffness estimation and
relative stiffness estimation. Absolute stiffness estimation
aims to determine the physical stiffness parameter, typically
defined as the ratio between applied force and resulting
deformation according to Hooke’s law. Traditional methods
rely on force–displacement measurements obtained from
sensors (e.g., tactile sensor, force–torque sensor, or inden-
tation devices), where controlled probing actions are applied
to derive the stiffness \cite{wang2016overview}. More recently, learning-based techniques have been explored to infer stiffness from multimodal sensory data such as force signals, tactile images, and deformation measurements \cite{hattori2015contact,gao2016deep}. For example, Bednarek et al.~\cite{bednarek2021gaining} proposed a deep learning framework that estimates stiffness during squeezing using IMU signals collected from the soft gripper.

In order to achieve stiffness estimation and adpative grasping, traditional rigid grippers typically employ multiple rigid fingers to enable accurate position and force control \cite{he2023rigid}. However, safely interacting with objects with varying mechanical properties requires compliant mechanisms and reliable force estimation capabilities. To address this challenge, Ruiz et al.~\cite{ruiz2022compliant} proposed a compliant gripper with four underactuated fingers capable of estimating grasping forces using a kinetostatic finger model, enabling safe physical human–robot interaction. Beyond specialized hardware, recent work has explored leveraging intrinsic sensing within robotic grippers to infer contact conditions without external sensors. For instance, Cortinovis et al.~\cite{cortinovis2025stiffness} developed algorithms that detect finger–object collisions and estimate stiffness using intrinsic sensors embedded in standard grippers.

In parallel, bio-inspired soft grippers have been widely studied for manipulating delicate and deformable objects due to their inherent compliance and adaptability~\cite{odhner2013open,li2019vacuum,manti2015bioinspired,shintake2018soft}. Soft robots can further enhance durability and robustness through mechanisms such as self-healing pneumatic structures \cite{terryn2017self}. Physics-based modeling approaches have also been explored to estimate contact forces and stiffness in soft robotic grippers. For example, Ghanizadeh et al.~\cite{ghanizadeh2024contact} employed finite element modeling combined with experimental validation to predict contact forces and enable closed-loop force control. Similarly, Herrera et al.~\cite{herrera2025physically} introduced a soft haptic whisker capable of estimating object stiffness through deformation modeling based on Cosserat rod theory while maintaining very low interaction forces.

Due to the inherent compliance and sensing capabilities of pneumatic actuators, pneumatic grippers provide a promising platform for stiffness estimation. In such systems, the relationship between internal pressure, actuator deformation, and object reaction forces can be exploited to infer object stiffness. A stiffness sensing framework using a soft pneumatic gripper performing a pincer grasp was present in ~\cite{sithiwichankit2023advanced}, where actuator deformation and pressure signals are used to estimate object stiffness. Similarly, Jang et al.~\cite{jang2024soft} investigated the use of pressure and deformation measurements in pneumatic fingers to infer contact forces and grasp states. Pneumatic actuation has also been utilize to enable variable stiffness manipulation; for example, modulating internal pressure allows soft grippers to adapt their stiffness for different manipulation tasks was demonstrated in ~\cite{song2025soft}. In addition, Grady et al.~\cite{grady2022visual} proposed a visual pressure estimation and control method that infers the pressure applied by a soft gripper using RGB images from an external camera.

These previous studies demonstrate the growing interest in stiffness estimation of soft gripper. In particular, pressure-based sensing in pneumatic grippers provides an effective approach for estimating object stiffness and enabling adaptive grasping.

\section{SELF-SENSING HYBRID GRIPPER DESIGN}
\begin{figure}[t]
    \centering
    \includegraphics[width=1\linewidth]{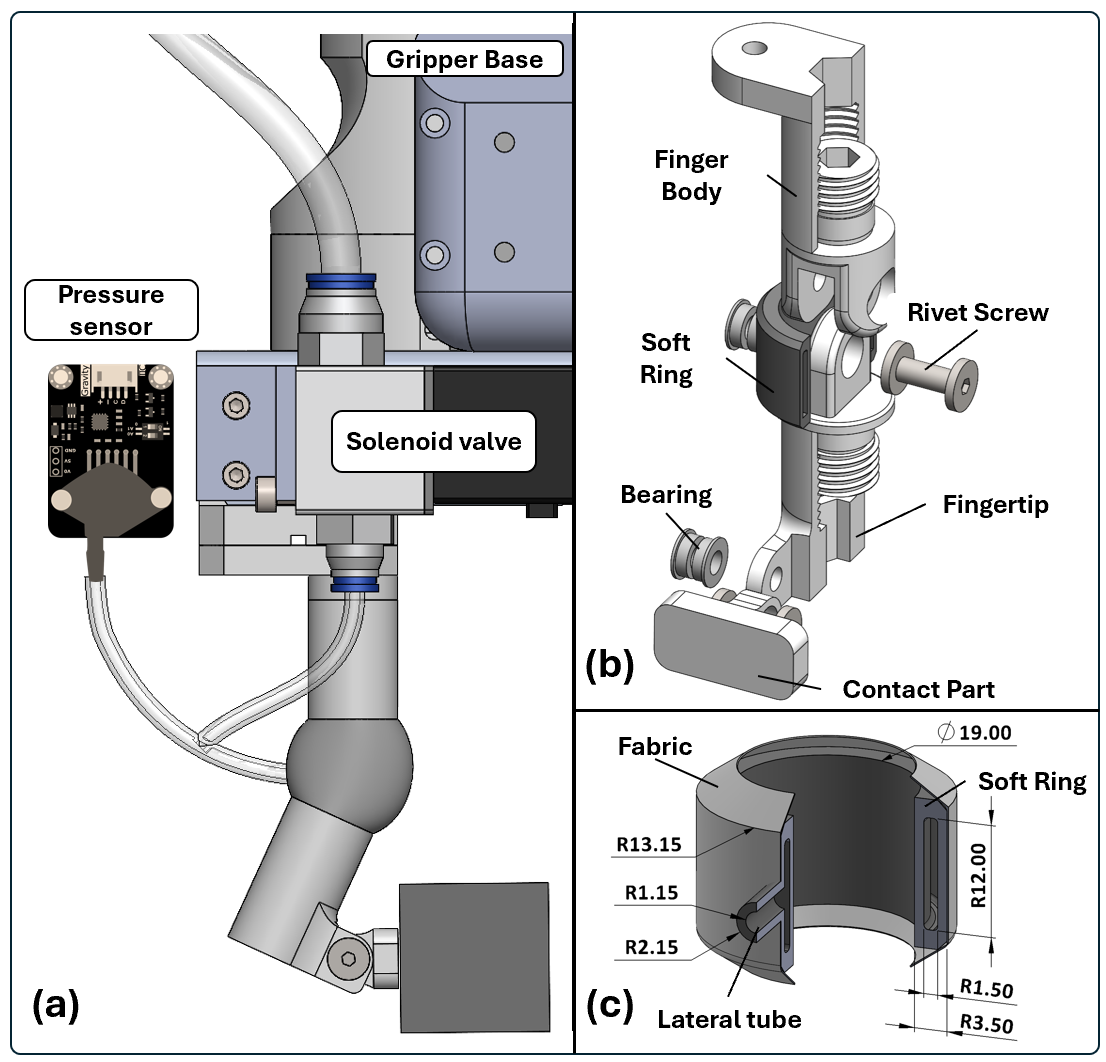}
    \caption{\textbf{Hardware Details:} Structural components and dimensions of the newly designed finger and soft pneumatic system.
(a) The pneumatic system consists of a pressure sensor and a soft ring, both controlled through the same air channel connected to a solenoid valve.
(b) Cross-sectional view of the hybrid finger showing the newly designed fabric-attaching thread mechanism integrated into the finger body.
(c) The drawing illustrates the cross-sectional dimensions of the soft core, which includes an additional tube designed to simplify the installation of the silicone tube.}
    \label{fig: Hardware Details}
\end{figure}

The proposed self-sensing hybrid gripper consists of three main components: (1) a rigid gripper finger (Fig.~\ref{fig: Hardware Details}(b)), (2) a soft ring for air-pressure sensing (Fig.~\ref{fig: Hardware Details}(c)), and (3) an air supply and regulation system (Fig.~\ref{fig: Hardware Details}(a)). The design details of each subsystem are presented in the following sections.

\label{sec:design}

\subsection{Gripper finger design}
The design of this soft-rigid hybrid finger is mainly motivated by the hybrid gripper proposed for grasping paper~\cite{Tran2025}. The rigid components of the finger, consisting of two links connected by a rotational joint, are comparable in size to an average human finger, with a total length of \SI{80}{\milli\meter}. Two flange bearings and a rivet screw are integrated into the joint to enable passive backward bending (as shown in Fig.~\ref{fig: Hardware Details}(a)).

The fingertip can bend up to \SI{80}{\degree} to facilitate object manipulation. At the bottom of the fingertip, it features a rotatable structure that allows for maintaining consistency in the direction of the normal force applied to the object. Friction within this joint of the structure is minimized by incorporating two bearings and a rivet screw. Both the finger body and the fingertip feature a specialized threading mechanism, shown in Fig.~\ref{fig: Hardware Details}(b), that securely anchors the soft ring’s fabric layer, enhancing the structural integrity of the hybrid finger.


\subsection{Soft ring design}
The soft component of the hybrid gripper, designed in the shape of a ring, provides both adaptability and self-sensing capability when incorporated with a pressure sensor. The dimensions of the soft ring are consistent with those demonstrated in~\cite{Tran2025}, covering the entire finger joint, with an inner diameter of \SI{17}{\milli\meter} and an outer diameter of \SI{20}{\milli\meter}. We added a small lateral soft tube to accommodate a \SI{4}{\milli\meter} diameter silicone tube that supplies compressed air to the ring. This additional tube simplifies installation and reduces the risk of air leakage.

The soft ring (as shown in Fig.~\ref{fig: Hardware Details}(c)) is 3D printed using a Formlabs 3B+ printer with Silicone 40A (Shore A hardness 40). The cavity inside the soft ring is connected with a silicone tube, sealed by Shin-Etsu silicone sealant. A thin layer of non-stretchable chiffon fabric is wrapped and bonded around the soft ring to concentrate the pneumatic pressure onto the finger structure.

\subsection{Gripper's self-sensing pneumatic system}

The self-sensing system of the hybrid gripper consists of soft rings that function as actuators, allowing the generation of the grasping force by a passive bending fingertip. With the additional installation of a pressure sensor, the gripper can generate grasping force through pneumatic actuation of the soft rings while simultaneously estimating the object’s stiffness by monitoring internal pressure variations during the probing process.



For pressure measurement within the soft ring, an MPX5700AP (DFRobot) pressure sensor module is connected to the ring through an additional air channel arranged in parallel with the soft ring. The sensor is capable of measuring a wide pressure range and has a maximum measurement deviation of approximately 2.5\%.

To separately accommodate volume variations during the operation of the two hybrid fingers, two solenoid valves are installed upstream of both the soft ring and the pressure sensor within the air channel. These valves isolate and lock the air inside the system, enabling accurate measurement of volume changes.

\begin{figure}
    \centering
    \includegraphics[width=1\linewidth]{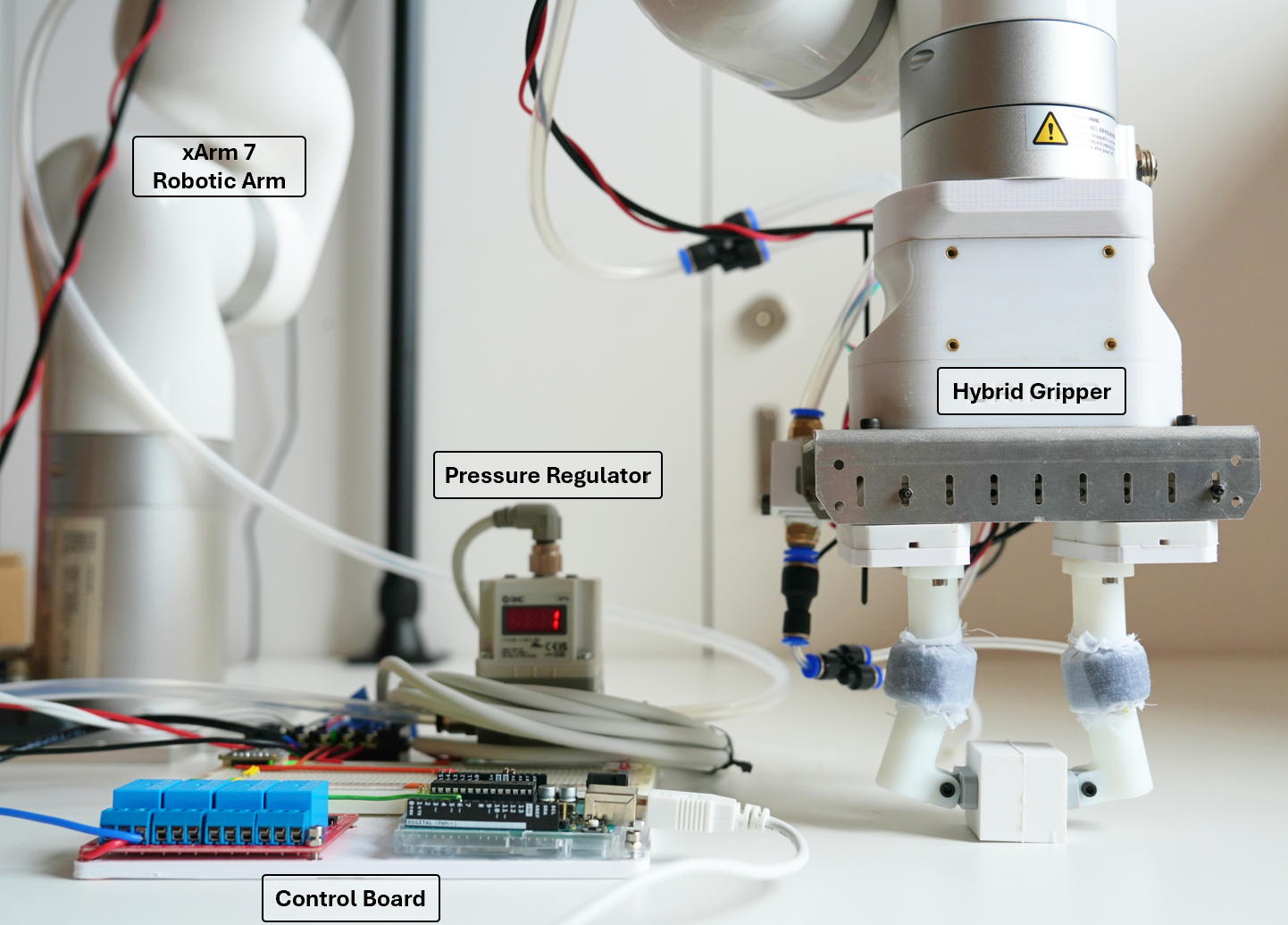}
    \caption{\textbf{System hardware:} Experimental platform comprising an xArm 7 robotic arm equipped with a hybrid gripper, a pneumatic supply line with a pressure regulator (SMC CORPORATION), and a control board responsible for command generation, signal interfacing, and overall system integration.}
    \label{fig: System Hardware}
    \vspace*{\duy}
\end{figure}

\subsection{Hardware system, gripper prototype and robot arm}
After the fabrication of the soft ring and rigid components is completed, the hybrid fingers are installed into the gripper base. The gripper base is closed-loop controlled by a NEMA17 stepper motor paired with an MKS SERVO42C motor driver, enabling linear inward and outward movement through a solid metal frame.

To facilitate the grasping and probing procedures, the entire hybrid gripper is mounted on a 6-DOF commercial xArm7 robotic arm. This configuration provides the gripper with a high degree of mobility and positional control, allowing it to approach objects from different orientations and perform controlled manipulation tasks. As a result, consistent grasping and probing actions can be executed across different trials, ensuring reliable data collection and improving the overall accuracy of the experimental evaluation conducted in this research.

Regarding the control system, the hybrid gripper is operated by an integrated controller board. The board incorporates a pressure regulator controlled by an MCP4725 12-bit DAC for pneumatic regulation via an Arduino Uno. In addition, an Arduino Nano RP2040 manages the overall system control functions and records data from the pressure sensor.

\section{EXPERIMENTS} \label{sec:experiment}

\subsection{Soft joint stiffness experiments}
\subsubsection{Soft joint stiffness experiments conducted without air volume constraints}\label{sec: Soft joint stiffness experiments conducted without air volume constraints}
\begin{figure}
    \centering
    \includegraphics[width=1\linewidth]{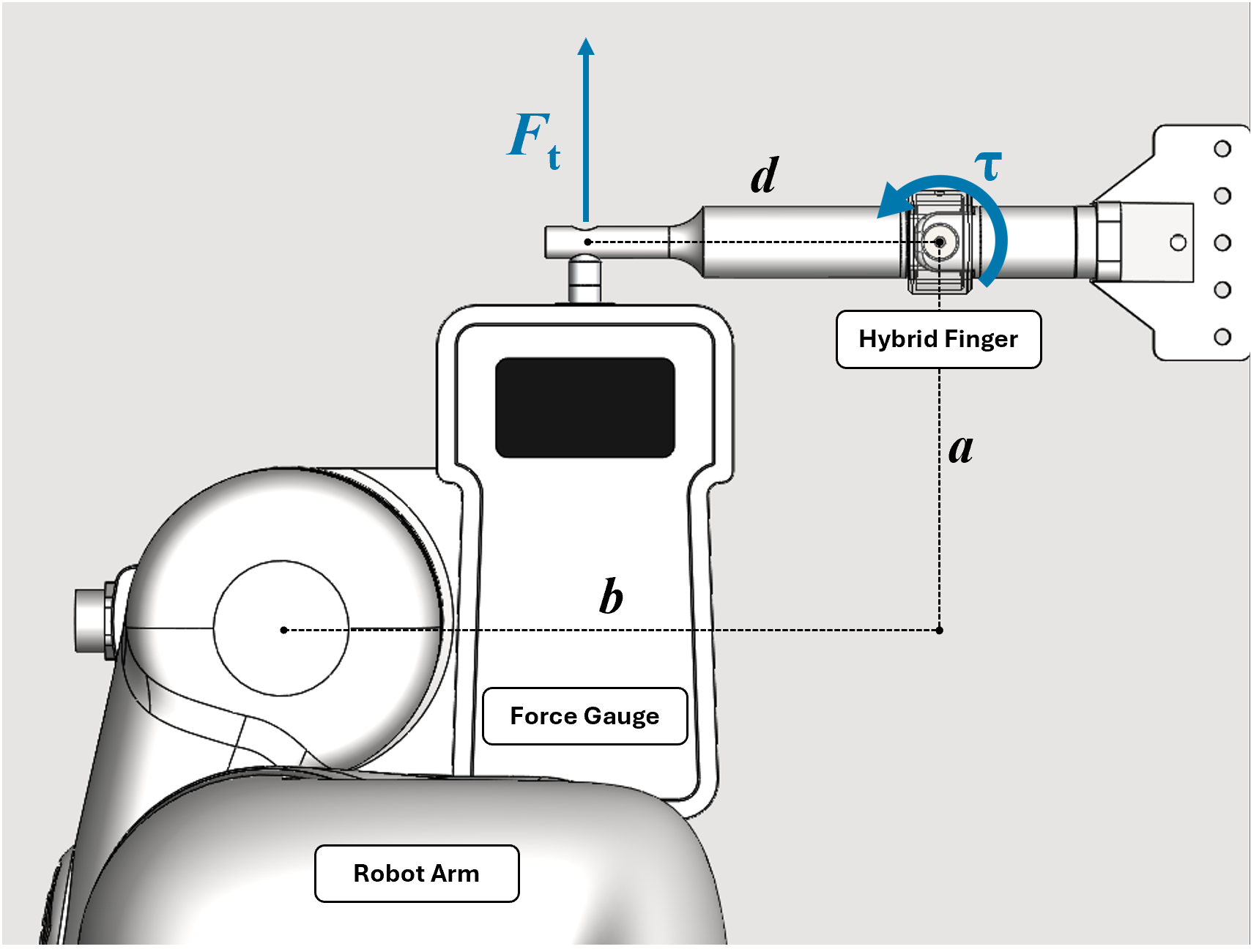}
    \caption{Demonstration of the data-collection procedure used to validate stiffness estimation. A force gauge mounted on the robot arm follows a circular trajectory around the joint as the fingertip bends. For each bending angle, the torque $\uptau$ generated by the soft ring is computed from the tangential force $F_t$ and the moment arm length.}
    \label{fig: experiment 1}
    \vspace*{-5mm}
\end{figure}

The first experiment is conducted to determine the relationship between the bending angle, compressed air pressure, and the torque exerted at the fingertip. In this experiment, the finger is equipped with a pneumatic system and fixed horizontally on the experimental base using a mounting fixture. An Nidec FGP-0.5N force gauge is mounted on a 6-DOF xArm7 robotic arm to generate circular motion in the horizontal plane around the rotation axis of the hybrid finger. The position of the end effector, represented by $x_{e}$, $y_{e}$, and $yaw_{e}$ in the $O_{xy}$ plane, can be expressed as a bending angle $\alpha$ about the rotation axis $O$, located at $(x_o, y_o)$, shown in Fig.~\ref{fig: experiment 1}.

\begin{equation}
\begin{gathered}
x_{e} = x_{o} + \sqrt{a^2 + b^2}\sin(\alpha - \beta) \\
y_{e} = y_{o} + \sqrt{a^2 + b^2}\cos(\alpha - \beta) \\
yaw_{e} = yaw_o - \alpha
\end{gathered}
\label{eq:torque_experiment_1}
\end{equation}

with $a$ and $b$ are the designed dimension parameter length of the hybrid finger, $\alpha$ is the bending angle of the fingertip, and $\beta$ is the design angle of the finger, as shown in Fig.~\ref{fig: Hardware Details} and Fig.~\ref{fig: experiment 1}.
    
For bending angles from \SI{0}{\degree} to \SI{80}{\degree} with \SI{5}{\degree} increments, and for compressed-air pressure supplied to the soft ring from 0 to \SI{150}{\kilo\pascal} in \SI{5}{\kilo\pascal} increments, the normal force $F_n$ is recorded using a force gauge at each bending angle, as shown in Fig.~\ref{fig: experiment 1}. Finally, the torque is calculated using the following equation:
    
    \begin{equation}
    \uptau = F_td
    \label{eq: torque experiment 1}
    \end{equation}
    where $\uptau$ denotes the torque generated by the soft ring at the fingertip, and $d$ represents the length of the experimental fingertip arm. Consequently, the relationship among these three parameters, bending angle, the pressure supplied to the soft ring, and the resulting torque, is illustrated using a heat map in Section~\ref{sec: result, ex1.1}

\subsubsection{Soft joint stiffness experiments with air volume constraints}

With the same method conducted in experiment~\ref{sec: Soft joint stiffness experiments conducted without air volume constraints}, the relationship between the change of pressure with a fixed amount of air inside the soft ring and the torque generated by the soft ring, paralleled with the bending of the fingertip, is validated. 

To completely validate this relationship, for each pressure inside the range from 0 to \SI{100}{\kilo\pascal}, both the change in pressure and the torque will be measured according to the bending angle of the fingertip in the same setup in experiment~\ref{sec: Soft joint stiffness experiments conducted without air volume constraints}. When the pressure is set to a specific value, the solenoid valve locks the current amount of air inside the ring. With the rotation movement of the force gauge with an angle increment of \SI{1}{\degree} around the axis of the finger joint, the force and the change in pressure would be measured by the force gauge and pressure sensor installed in a parallel air channel with a soft ring.

The reason behind this change in pressure corresponds to the change in the volume of the soft ring. When the bending angle increases, the fingertip will occupy the space, leading to a decrease in the volume inside the soft ring. In addition, the data when the fingertip moves back to the initial position is also recorded to validate the hysteresis characteristic of the hybrid system.

\subsection{Proposed probing method to estimate object stiffness}\label{sec: experiment, probing method}
The primary objective of this method is to estimate the normal force generated when the fingers close inward by a specific distance $d_\mathrm{c}$ through the soft ring. Subsequently, the relative stiffness of the object is determined using the standard stiffness equation based on the estimated normal force.
\begin{equation}
    k_\mathrm{r}=\frac{F}{d_\mathrm{c}}
\label{eq: object stiffness reference}
\end{equation}
where $F$ is the force estimated by the change of pressure inside the soft ring when contact happens, and $d_\mathrm{c}$ is the constant closing distance of the hybrid finger.

Furthermore, the deformation of the object along the x-axis can be calculated using the following equation for the given configuration of the hybrid gripper, corresponding to the bending angle $\alpha$: 
 \begin{equation}
    \delta = d_\mathrm{c}-(a+ \sqrt{a^2+b^2}\sin{(\alpha - \beta)})
\label{eq: object deformation}
\end{equation}



The probing process consists of several sequential steps. First, the gripper is fully opened to its maximum range. It is then slowly closed toward the object until initial contact occurs. The contact position is identified by detecting a change in the pressure sensor signal at the moment of contact.


After detecting the initial contact, the gripper advances toward the object with a predefined incremental displacement for five probing times. During each probing time, the change of the pressure is measured with the strategies of controlling the initial pressure. These measurements are used to normal contact force $F_n$. Finally, based on the constant closing displacement of the finger and normal force, the relative stiffness of the objects can be determined and compared with each other.





\subsection{Object stiffness validation experiments}
This experiment is conducted to validate the stiffness discrimination capability based on the ground truth value of several objects.



To evaluate the stiffness discrimination capability, there are three different cubes with different stiffness would be probed by the hybrid gripper. The ground-truth stiffness value of these object are obtained by a system of a robot arm and a scale, which robot arm creates the precise downward movement for the probing head, and the scale measures the normal force on the cube. The calculated stiffness for objects is presented in the Table~\ref{tab: relative_stiffness}. The cubes were fabricated from different materials: Cube 1 was made from Shin-Etsu KE-1415 with a 3\% mixing ratio, Cube 2 was made from Shin-Etsu KE-1415 silicone with a 5\% mixing ratio, and Cube 3 was made from Flexible 80A material.

\begin{table}[h]
\centering
\caption{Relative stiffness of tested objects}
\label{tab: relative_stiffness}
\begin{tabular}{lc}
\hline
\textbf{Objects} & \textbf{Relative Stiffness (N/mm)}\\
\hline
Cube 1 (20A)  & 50.83 \\
Cube 2 (30A)   & 54.87 \\
Cube 3 (80A)   & 202.39 \\
\hline
\end{tabular}
\end{table}

\begin{figure*}
    \centering
    \includegraphics[width=1\linewidth]{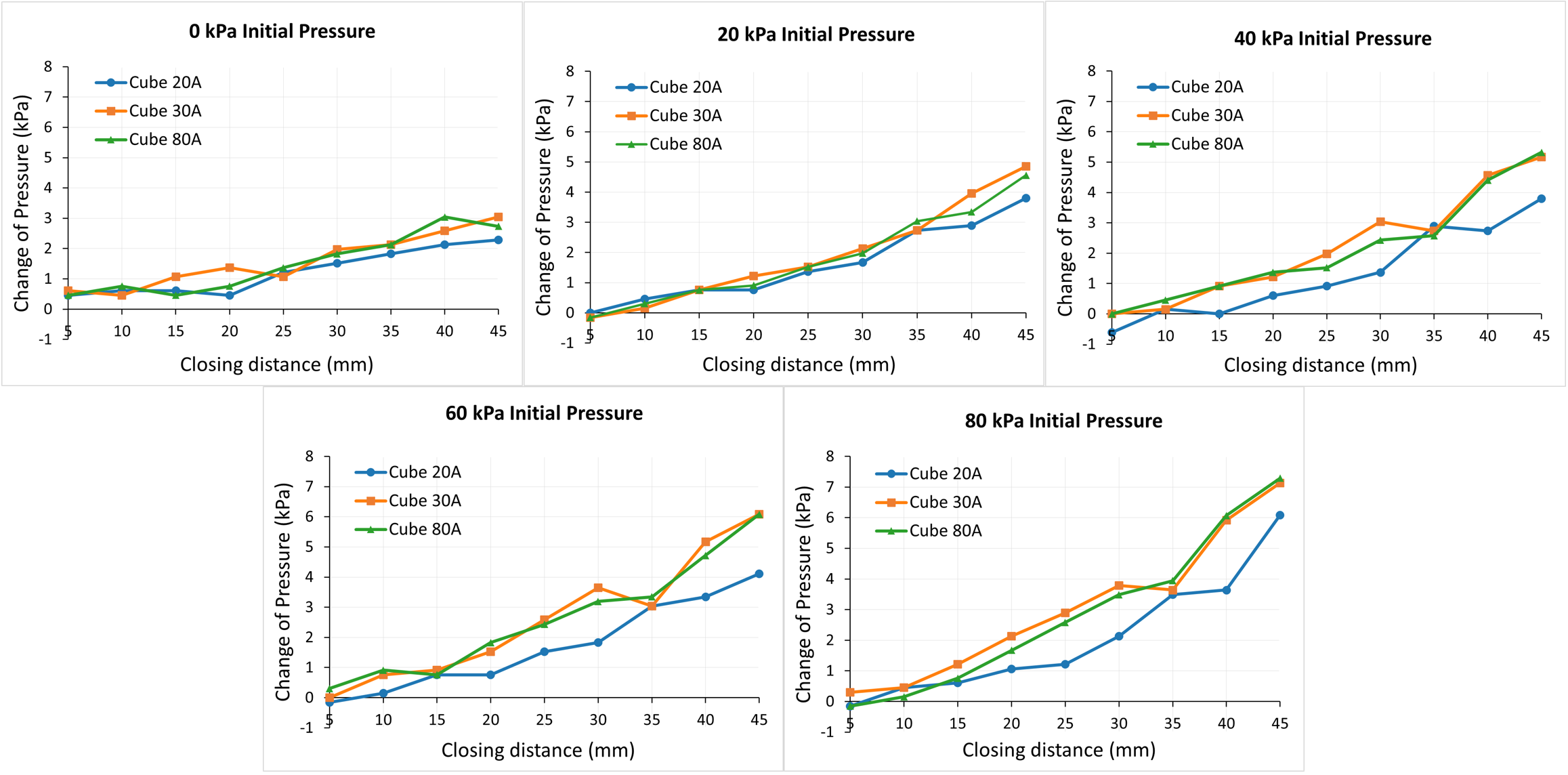}
    \caption{Pressure variation during the probing process under different initial pressures. Each subplot shows the relationship between the change in pressure and the gripper closing distance for objects with different stiffness levels (Cube 20A, Cube 30A, and Cube 80A). As the closing distance increases, the pressure change becomes more significant for stiffer objects, demonstrating that pressure variation can be used to distinguish relative object stiffness. In addition, the gripper demonstrates higher sensitivity in differentiating stiffness at an initial pressure of \SI{60}{\kilo\pascal} and a closing distance of approximately \SI{30}{\milli\metre}.}
    \label{fig: 5}
    \vspace*{\duy}
\end{figure*}

Afterward, the hybrid gripper sequentially applied the probing method to the sides of the cubes while varying the closing distance and the initial pressure supplied to the soft ring. In this experiment, five different initial pressure values were tested, ranging from \SI{0}{\kilo\pascal} to \SI{80}{\kilo\pascal} with increments of \SI{20}{\kilo\pascal}. These conditions are illustrated by five line graphs in Fig.~\ref{fig: 5}. For each initial pressure value, the gripper performed probing with several closing distances.

Based on the results of this experiment, the proposed probing method for distinguishing stiffness between objects using only pressure change measurements has been validated. In addition, the most sensitive pair of parameters (initial pressure and closing distance) for stiffness discrimination can be identified from this experiment, enabling improved performance in real-world grasping and probing tasks in the subsequent experiment.

\subsection{Stiffness probing and grasping experiments}
In this experiment, grasping combined with object stiffness estimation in real-world applications is evaluated through the proposed result obtained in the probing method.



A rooted orange or a ripe banana are choose as the experimental objects. These objects are placed firmly in the experiment's working space. These types of fruits typically exhibit varying levels of stiffness due to non-uniform ripening. To grasp the fruit in this scenario, the gripper must probe and identify areas that are firm enough for secure grasping while avoiding softer, riper regions that could be damaged during manipulation.

To address this issue, the fruit is first placed within the robot’s working space. The probing locations are then determined using a hybrid gripper and an appropriate probing strategy based on the fruit’s shape.

For the orange, the hybrid gripper mounted on a robotic arm rotates its wrist to probe the fruit from multiple angles, ranging from \SI{0}{\degree} to \SI{180}{\degree} in steps of \SI{30}{\degree}.

For the banana, the probing process is conducted along its elongated body at ten positions that are equally spaced along the $O_y$ axis of the system.

After collecting the pressure variation data and converting it into the relative stiffness of the object, the hybrid gripper selects the grasping location with the highest stiffness while avoiding softer regions that could damage the fruit.


\section{EXPERIMENT RESULTS}

\subsection{Soft joint stiffness experiments} \label{sec: result, ex1.1}
\subsubsection{Soft joint stiffness experiments conducted without air volume constraints}
\begin{figure}
    \centering
    \includegraphics[width=\linewidth]{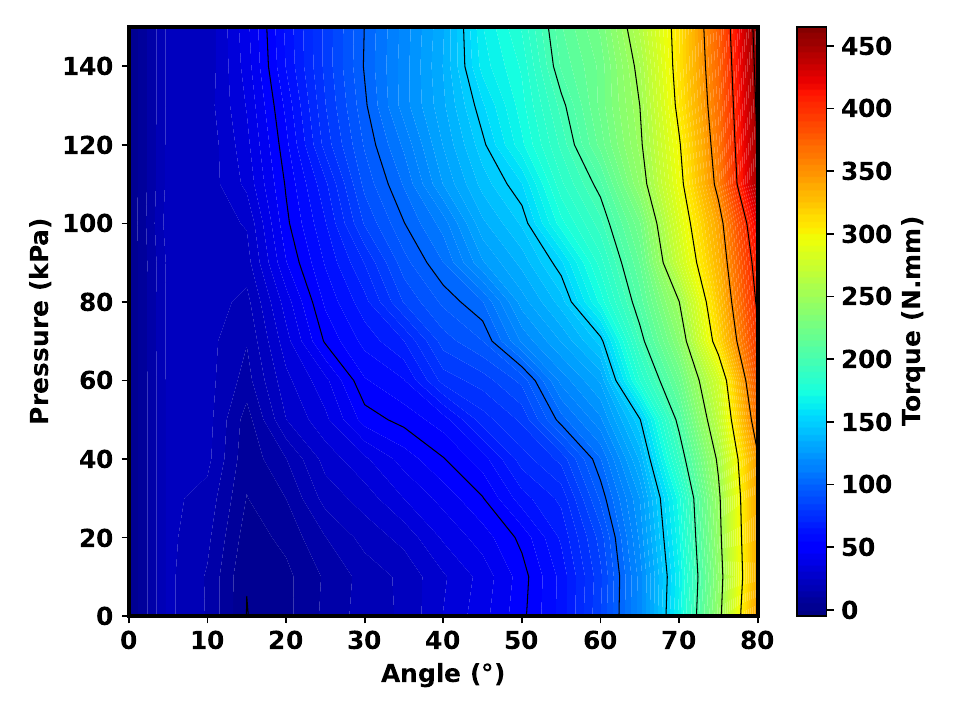}
    \caption{The torque was measured in soft-joint stiffness experiments, with pressure continuously controlled by a regulator from 0 to 150 kPa and bending angles ranging from \SI{0}{\degree} to \SI{60}{\degree}.}
    \label{fig: Experiment 1 Result}
    \vspace*{-5mm}
\end{figure}

 Using silicone 40A, which has higher Shore Hardness compared with the silicone used in research~\cite{Tran2025}, the stiffness of the joint in this research can go up to 450 kPa with the bending angle of \SI{80}{\degree}, as shown in Fig.~\ref{fig: Experiment 1 Result}. 

Regarding the bending angle range from \SI{0}{\degree} to \SI{20}{\degree}, the torque does not change significantly. This result may be attributed to the characteristics of the fabric layer, which loosens at the initial stage of bending and only begins to influence the torque after it is fully stretched beyond \SI{20}{\degree}.

\subsubsection{Soft joint stiffness experiments with air volume constraints}
\begin{figure}
    \centering
    \includegraphics[width=0.75\linewidth]{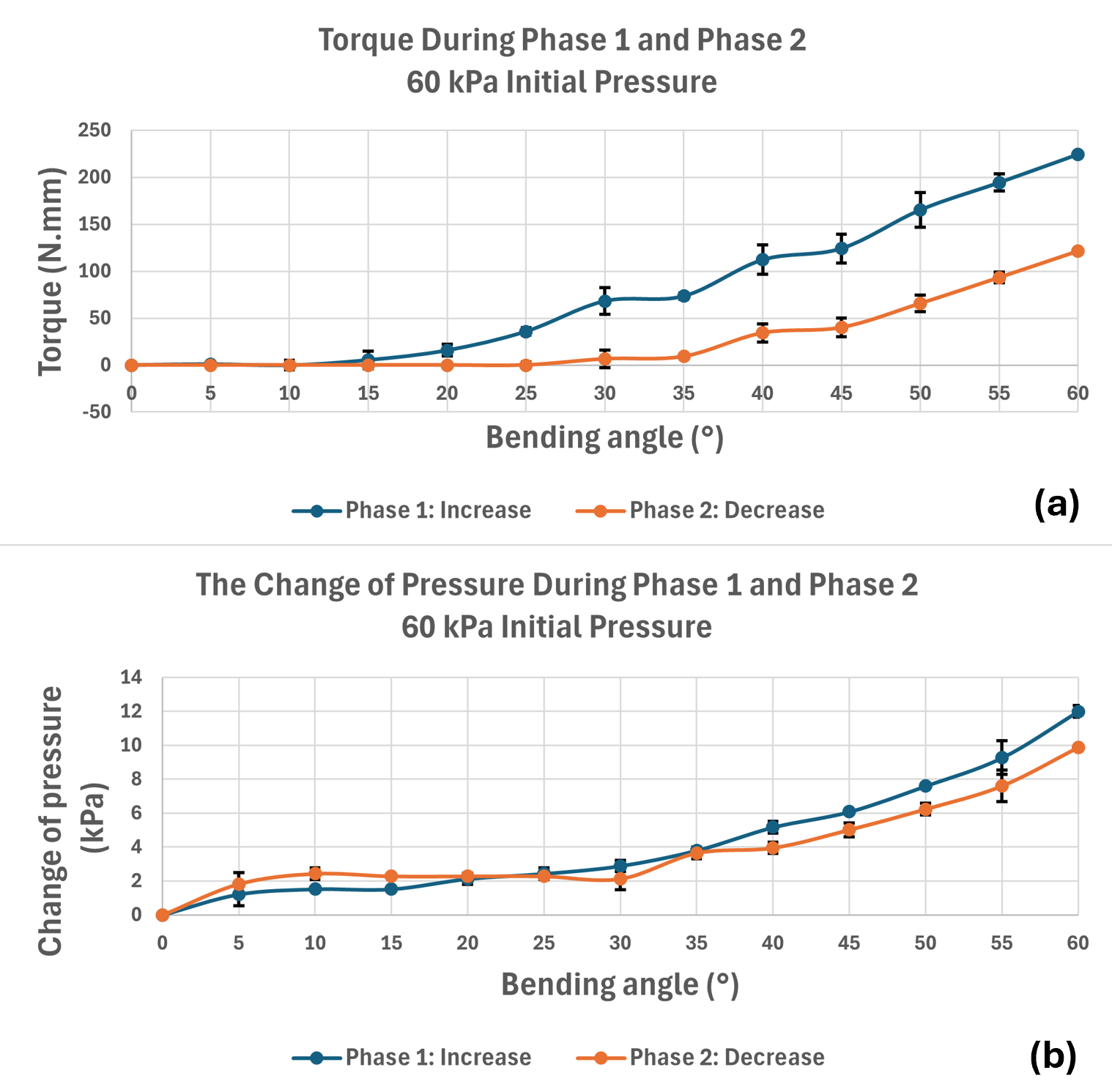}
    \caption{(a) Torque–bending angle relationship during Phase 1 (increasing) and Phase 2 (decreasing). (b) Corresponding pressure variation as a function of bending angle for both phases. In Phase 1, data were recorded as the fingertip bent progressively from \SI{0}{\degree} to a specified angle. In Phase 2, data were recorded as the fingertip bent to a maximum angle of \SI{80}{\degree} and then returned to the specified angle, where measurements were taken.}
    \label{fig: Experiment 2 Result}
    \vspace*{\duy}
\end{figure}

Fig.~\ref{fig: Experiment 2 Result} shows that when a constant amount of air is maintained inside the soft ring, the relationship among the torque generated by the soft ring, the initial pressure, and the bending angle is relatively similar to the results obtained in experiment~\ref{sec: result, ex1.1}. However, within the bending angle range of \SI{0}{\degree} to \SI{25}{\degree}, the looseness of the fabric layer significantly affects the torque generated by the soft ring, resulting in a torque value close to \SI{0}{\newton\milli\metre} in this range.

The linear relationship between the change in pressure and the bending angle is noticeable in Fig.~\ref{fig: Experiment 2 Result}(b). This result clearly validates the method used to determine the contact position during the probing process described in Section~\ref{sec: experiment, probing method}.

In addition, data from the reverse bending process of a specific initial pressure are also recorded in both graphs. The difference between the two colored lines, shown in Fig.~\ref{fig: Experiment 2 Result} and representing the forward and backward bending directions, can be attributed to leakage at the fitting joints in the pneumatic system. Such leakage affects the pressure variation when the solenoid valve attempts to maintain a constant amount of air inside the soft ring.
\begin{figure}
    \centering
    \includegraphics[width=1\linewidth]{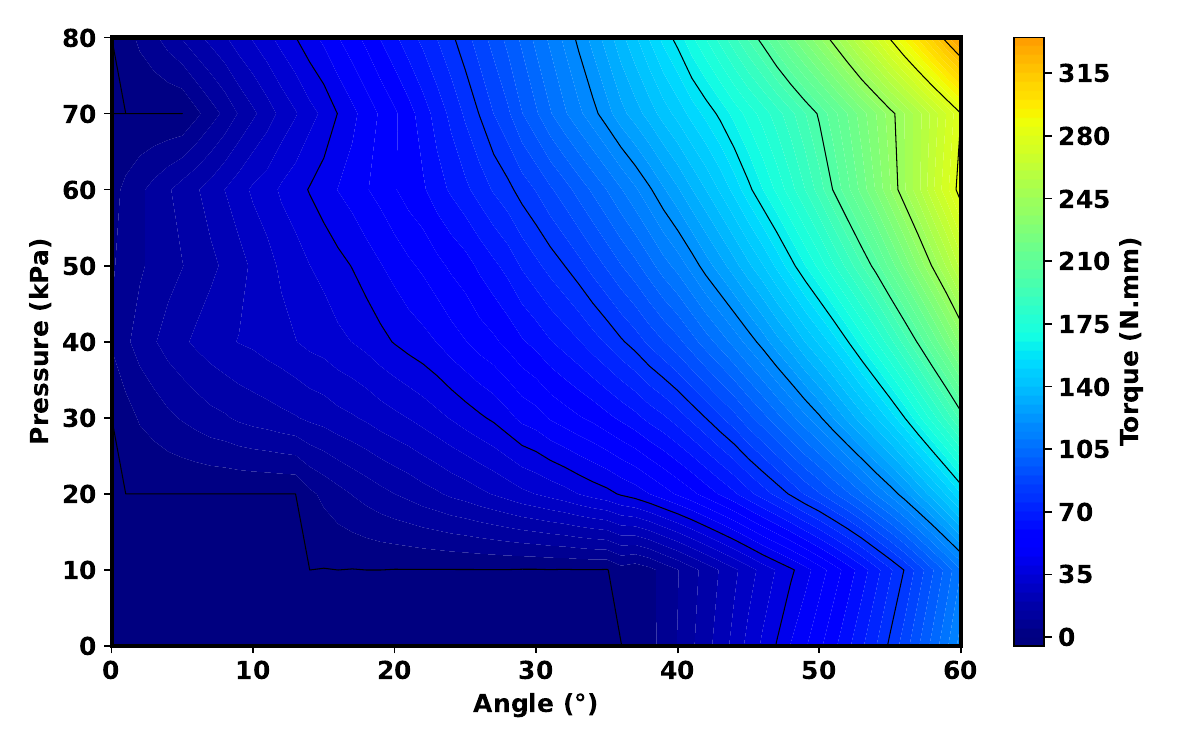}
    \caption{The torque was measured in soft-joint stiffness experiments under constrained air-volume conditions by solenoid valves, with bending angles ranging from \SI{0}{\degree} to \SI{60}{\degree} and pressures ranging from 0 to 80 kPa.}
        \label{fig: Experiment 1 result: fix_pressure_angle_torque}
        \vspace*{\duy}
\end{figure}

\subsection{Object stiffness validation experiments}

Hooke’s law is used to obtain the general stiffness of the object:
\begin{equation}
    k_\mathrm{o}=\frac{F}{\delta}
\label{eq: object stiffness reference2}
\end{equation}

 From Fig.~\ref{fig: Experiment 1 result: fix_pressure_angle_torque}, we observe that the bending angle is directly proportional to both the pressure change and the torque generated by the soft ring. In addition, the relative deformation of the object is inversely proportional to the bending angle of the fingertip after contact happens, as implied by equation~\eqref{eq: object deformation}. Therefore, based on the formula for the relative stiffness of the object presented in equation~\eqref{eq: object stiffness reference}, we can conclude that the relative stiffness of the object is directly proportional to the pressure change of the soft ring. This result indicates that measuring the pressure change after probing an object with this hybrid gripper is sufficient to distinguish the relative stiffness between different objects.

\subsection{Stiffness probing and grasping experiments}
In this experiment, the proposed probing method is applied to measure pressure changes in order to infer the relative stiffness at defined positions on the bodies of a banana and an orange.

\begin{figure}[h]
    \centering
    \includegraphics[width=0.85\linewidth]{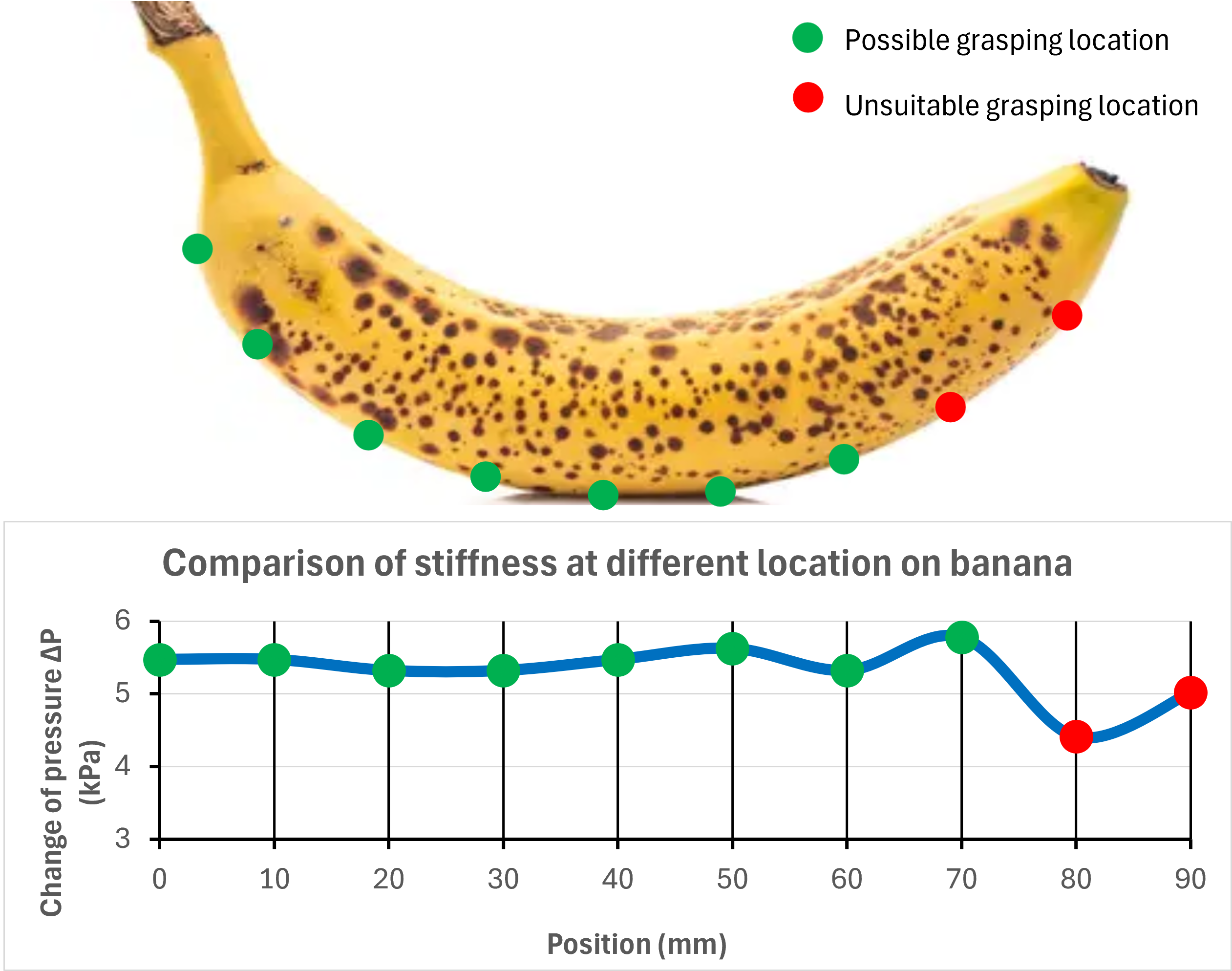}
    \caption{The relative stiffness of the banana was measured at different positions along its length. The softest region coincided with the black spot on the peel.}
    \label{fig:banana}
\end{figure}

Fig.~\ref{fig:banana} indicates the variation in pressure change along the body of the banana. From position 0 to \SI{70}{\milli\metre}, the pressure change is not noticeable. However, at positions 80 mm and 90 mm, denoted by red marks, the pressure change is significantly lower, indicating that the stiffness at these positions is lower than at the others. Similar to the banana probing process, the hybrid gripper identifies locations with lower pressure changes after rotating to probe around the orange at \SI{30}{\degree}, \SI{60}{\degree}, and \SI{90}{\degree}. Based on these data, the possible grasping locations for both fruits are indicated by green marks, where the stiffness is estimated to be higher than at the locations marked in red.

\begin{figure}[h!]
    \centering
    \includegraphics[width=0.85\linewidth]{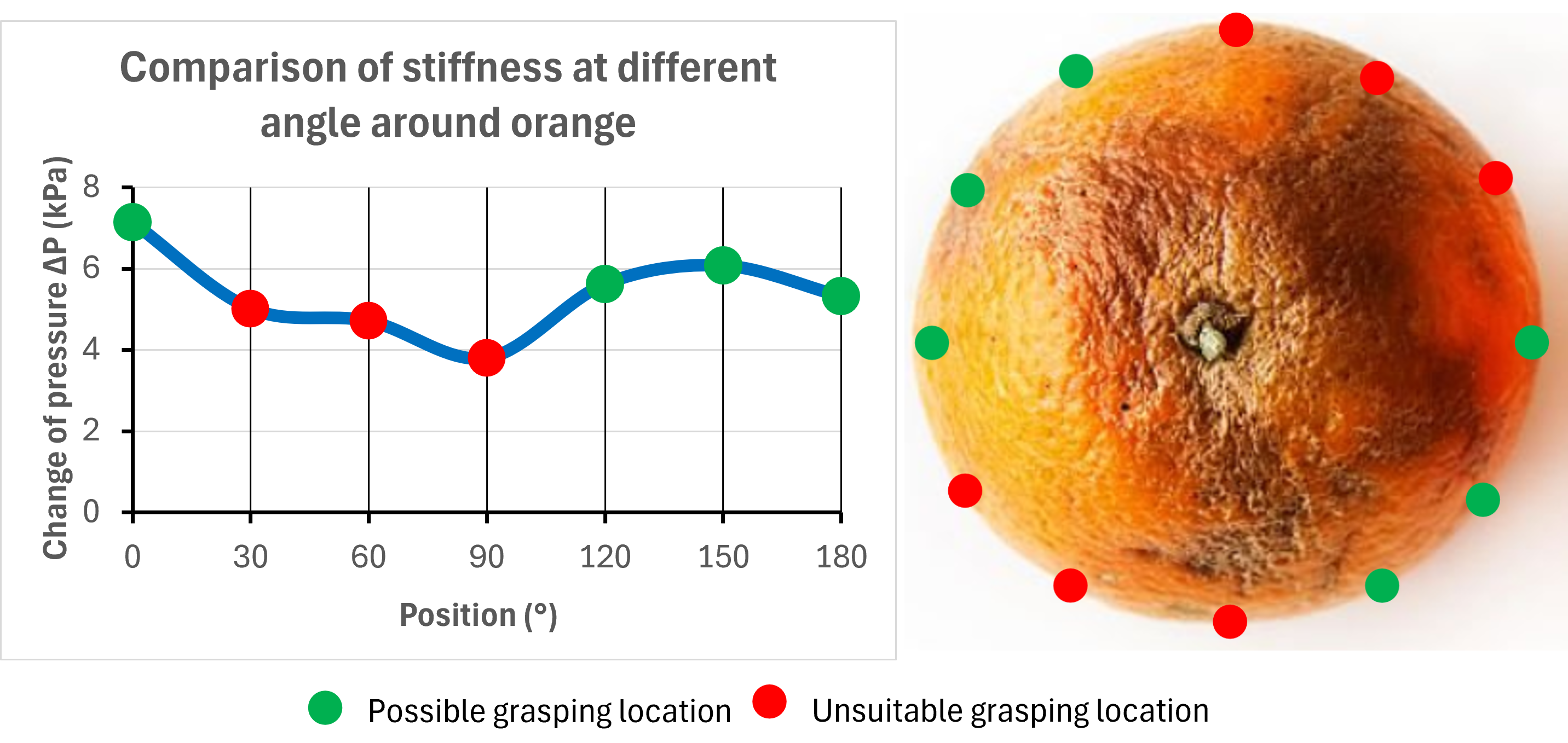}
                \caption{The relative stiffness of the semi-rooted orange was measured at different positions around its center. Based on these results, the orange should be grasped at the position with the highest stiffness to ensure a stable grip.}
    \label{fig:orange}
    \vspace*{\duy}
\end{figure}

\section{LIMITATIONS \& FUTURE WORK}
Although the proposed hybrid gripper demonstrates the capability to estimate the relative stiffness of objects through pressure variation, several limitations remain in the current system. First, the pressure sensor used in the system has relatively low resolution, which limits the accuracy of pressure change measurements during the probing process. This limitation could be addressed in future work by integrating a higher-resolution pressure sensor. In addition, hysteresis effects in the soft material, as well as unavoidable air leakage at the pneumatic fitting joints, may affect the repeatability and consistency of the measurements.
Furthermore, the pressure sensor is currently installed on only one fingertip of the gripper, which restricts the sensing capability during multi-contact interactions.

Future work will focus on addressing these limitations and improving the overall performance of the system. In particular, simulation software will be employed to thoroughly validate the proposed sensing principle and to analyze the deformation behavior of the soft ring under different loading conditions. These simulations will facilitate optimization of the structural design and sensing parameters of the hybrid gripper, thereby improving the accuracy, sensitivity, and robustness of stiffness estimation.

\section{CONCLUSIONS} \label{sec:conclusion}
This paper presented a probe-to-grasp manipulation framework built around a passive pneumatic hybrid gripper with self-sensing, variable-stiffness joints. By integrating rigid finger links with a soft pneumatic ring at each joint, the gripper achieves compliant interaction while allowing joint stiffness to be tuned through internal pressure. Instead of relying on embedded tactile or force sensors, we leveraged a single pressure sensor together with an experimentally identified lookup relationship between joint bending, pressure variation under constrained air volume, and the resulting joint torque. This enables practical estimation of contact forces during probing and grasping using minimal hardware. While the proposed approach offers a low-cost and easily deployable alternative to traditional tactile sensing, accuracy and repeatability can be further improved by addressing pressure sensor resolution. Future work will also include simulation-based analysis of soft ring deformation to optimize structural and sensing parameters and improve robustness across a broader range of objects and contact conditions.

\addtolength{\textheight}{-13cm}   

\bibliographystyle{IEEEtran}
\bibliography{biblio}
\end{document}